\documentclass[lettersize,journal]{IEEEtran}
\usepackage{amsmath,amsfonts}
\usepackage{algorithmic}
\usepackage{algorithm}
\usepackage{array}
\usepackage[caption=false,font=normalsize,labelfont=sf,textfont=sf]{subfig}
\usepackage{textcomp}
\usepackage{stfloats}
\usepackage{url}
\usepackage{verbatim}
\usepackage{graphicx}
\usepackage{cite}
\usepackage[pagebackref=true,breaklinks=true,colorlinks,bookmarks=false]{hyperref}

\usepackage{bbding}

\newcommand{\etal}{\textit{et al}.}

\newcommand{\eg}[1]{{\textit{e.g.}{{#1}}}}

\begin{document}

\title{Multi-Granularity Denoising and Bidirectional Alignment for Weakly Supervised Semantic Segmentation}

\author{Tao Chen, Yazhou Yao* and Jinhui Tang
		\thanks{Tao Chen, Yazhou Yao and Jinhui Tang are with the School of Computer Science and Engineering, Nanjing University of Science and Technology, Nanjing 210094, China.}
		\thanks{\copyright~2023 IEEE. Personal use of this material is permitted. Permission from IEEE must be obtained for all other uses, in any current or future media, including
			reprinting/republishing this material for advertising or promotional purposes, creating new collective works, for resale or redistribution to servers or lists, or reuse of any copyrighted component of this work in other works.}
	}

\markboth{}%
{Shell \MakeLowercase{\textit{et al.}}: A Sample Article Using IEEEtran.cls for IEEE Journals}


\maketitle

\begin{abstract}
Weakly supervised semantic segmentation (WSSS) models relying on class activation maps (CAMs) have achieved desirable performance comparing to the non-CAMs-based counterparts. However, to guarantee WSSS task feasible, we need to generate pseudo labels by expanding the seeds from CAMs which is complex and time-consuming, thus hindering the design of efficient end-to-end (single-stage) WSSS approaches. To tackle the above dilemma, we resort to the off-the-shelf and readily accessible saliency maps for directly obtaining pseudo labels given the image-level class labels. Nevertheless, the salient regions may contain noisy labels and cannot seamlessly fit the target objects, and saliency maps can only be approximated as pseudo labels for simple images containing single-class objects. As such, the achieved segmentation model with these simple images cannot generalize well to the complex images containing multi-class objects. To this end, we propose an end-to-end multi-granularity denoising and bidirectional alignment (MDBA) model, to alleviate the noisy label and multi-class generalization issues. Specifically, we propose the online noise filtering and progressive noise detection modules to tackle image-level and pixel-level noise, respectively. Moreover, a bidirectional alignment mechanism is proposed to reduce the data distribution gap at both input and output space with simple-to-complex image synthesis and complex-to-simple adversarial learning. MDBA can reach the mIoU of 69.5\% and 70.2\% on validation and test sets for the PASCAL VOC 2012 dataset. The source codes and models have been made available at \url{https://github.com/NUST-Machine-Intelligence-Laboratory/MDBA}.
\end{abstract}


\section{Introduction}

\IEEEPARstart{S}{emantic} segmentation, aiming to label every image pixel, has recently achieved remarkable success and demonstrated great application potential in fields like automatic driving and medical analysis  \cite{ding2020semantic,he2021mgseg,kang2019random,long2015fully,borji2015salient,noh2015learning,badrinarayanan2017segnet}. However, collecting accurate pixel-level annotations for training fully-supervised semantic segmentation models is considerably labor-intensive and time-consuming. As such, many researchers have paid attention to weak labels, including image-level annotations \cite{zhou2021group,jing2019coarse,jiang2021layercam,kolesnikov2016seed,wei2016stc,hong2017weakly,huang2018weakly,ahn2018learning,wei2018revisiting,jiang2019integral}, scribbles \cite{lin2016scribblesup}, bounding boxes \cite{dai2015boxsup}, and points \cite{bearman2016s}. Among these forms of supervision, the image-level label is the cheapest one while annotating \cite{bearman2016s} and it can also be directly obtained from many public datasets collected for the image classification task, \eg, ImageNet \cite{deng2009imagenet}. Therefore, the image-level label has become the most widely studied weak tag. In this paper, our focus is also the weakly supervised semantic segmentation (WSSS) task with image-level labels.

\begin{figure*}
	\centering
	\includegraphics[width=1.0\linewidth]{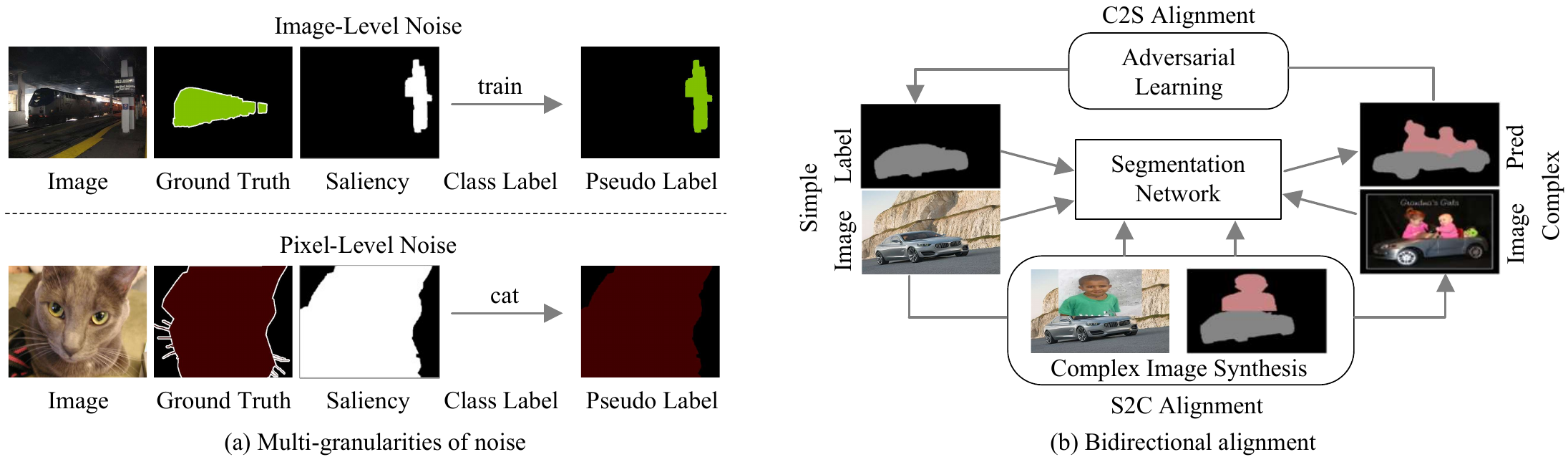}
	
	\caption{Motivation. (a) Multi-granularities of noise. With the help of image-level labels, we turn class-agnostic saliency maps into class-aware pseudo labels. However, the salient regions might not seamlessly fit the target objects or might even include other irrelevant non-target objects, leading to image-level and pixel-level noise. (b) Bidirectional alignment mechanism. Considering saliency maps can only be approximated as pseudo labels for simple images containing single-class objects, we propose a bidirectional alignment mechanism to reduce the data distribution gap at both input and output space with simple-to-complex (S2C) image synthesis and complex-to-simple (C2S) adversarial learning. }
	\label{fig_moti}
\end{figure*}

Recent WSSS methods typically train a classification network and rely on class activation maps (CAMs) \cite{zhou2016learning} to generate pseudo labels. However, CAMs can only locate the most discriminative parts of objects, bringing difficulty in obtaining accurate pixel-level pseudo labels for training the segmentation network. Therefore, recent works mainly focus on enlarging the activation area of CAMs to obtain more integral object regions. Unfortunately, the process of expanding segmentation seeds and generating reliable pseudo labels usually involves complex semantic propagation and/or time-consuming post-processing operations, which hinders the design of end-to-end WSSS approaches. Although some single-stage methods \cite{papandreou2015weakly,pinheiro2015image,hong2016learning,zhang2020reliability,zhang2021adaptive} have been explored recently, their performances are inferior compared to the contemporary multi-stage methods. Besides, existing single-stage methods usually need to simultaneously maintain a classification network for CAMs generation, a CAMs refinement module for better pseudo labels, and a target segmentation network. Consequently, these approaches rely heavily on many computing resources and tend to be less efficient.

Since it is difficult to leverage CAMs for designing efficient end-to-end (single-stage) WSSS models, in this paper, we resort to the off-the-shelf and readily accessible saliency maps \cite{hou2017deeply} for advancing single-stage WSSS task. Saliency maps are widely used in the multi-stage approaches \cite{wei2016stc,wei2018revisiting,jiang2019integral,fan2020learning,fan2020employing,zhang2020splitting,sun2020mining,yao2021non,lee2021railroad,wu2021embedded} and can provide useful boundary information. Inspired by \cite{wei2016stc} that leverages saliency maps for initial segmentation network training, we propose to directly train a single-stage WSSS model with saliency-generated pseudo labels. Specifically, as shown in Fig.~\ref{fig_moti}(a), we assign the image-level class label to the pixels belonging to the salient region, which transforms the class-agnostic saliency maps into pixel-level pseudo labels for the semantic segmentation task. However, the detected salient regions may have noisy locations, and thus cannot seamlessly fit the target objects or even can contain other irrelevant non-target objects (see the 3rd column of Fig.~\ref{fig_moti}(a)). Moreover, we find that saliency maps can only be well approximated as pseudo labels for simple images that contain one category of objects. For complex images with multiple classes of objects, we have no clue to assign different category labels to the class-agnostic salient regions. Due to the enormous distribution discrepancies between simple and complex images, training a segmentation network with only simple images cannot generalize well to the complex ones.

To tackle the above challenges, we propose a multi-granularity denoising and bidirectional alignment (MDBA) model to alleviate the noisy label issue and the distribution gap between simple and complex images. As shown in Fig~\ref{fig_moti}(a), we divide the types of noise into image-level and pixel-level noise according to the quality of pseudo labels, which is measured by the IoU between the pseudo label and the network prediction. Since the image-level noisy labels that contain too much noise may deteriorate the network training, we propose an online noise filtering module to directly discard them once detected during training, which can avoid misleading gradient back-propagation. Moreover, we propose a progressive noise detection module to identify wrongly labeled pixels with a gradually stricter loss threshold to deal with the pixel-level noisy labels. On the other hand, as shown in Fig~\ref{fig_moti}(b), to alleviate the multi-class generalization issue caused by training the model with only simple single-class images, we propose a bidirectional alignment mechanism to reduce the data distribution gap at both input and output space. Specifically, synthetic complex images are firstly generated from the simple ones for the simple-to-complex data distribution alignment at the input level. In addition, complex-to-simple alignment is proposed with adversarial learning at the output space, which aims to enforce the network predictions of complex images to be as close as possible to the pseudo ground truth of simple images. Extensive experiments on the PASCAL VOC 2012 and COCO datasets demonstrate the superiority of our proposed approach compared to existing single-stage methods. Treating MDBA as the pseudo label refinement step, our proposed method can be easily extended to a two-stage framework and achieve competitive results to current multi-stage approaches. To sum up, our contributions are:

\textbf{$\bullet$} We propose to solve the end-to-end WSSS task from the new perspective of leveraging non-CAMs-based pseudo labels.

\textbf{$\bullet$} We propose an online noise filtering module with class adaptive thresholds and a progressive noise detection module to tackle image-level and pixel-level noise.

\textbf{$\bullet$} We propose a bidirectional alignment mechanism to reduce the data distribution gap at both input and output space with simple-to-complex image synthesis and complex-to-simple adversarial learning.

\textbf{$\bullet$} Extensive experiments and ablation studies on PASCAL VOC 2012 and COCO datasets demonstrate the superiority of our approach and its complementarity with popular CAMs-based WSSS methods.

The rest of this paper is organized as follows: 
the related work is described in Section \ref{related_work} and our approach is introduced in Section \ref{approach}; we then report our evaluations and ablation studies on two widely-used datasets in Section \ref{experiments} and finally conclude our work in Section \ref{conclusion}.

\section{Related Work}
\label{related_work}

\subsection{Weakly Supervised Semantic Segmentation}
Though deep learning has recently achieved great success in semantic segmentation, training deep CNNs typically requires large-scale labeled datasets. Considering the difficulty in obtaining accurate pixel-level labels, researchers attempt to tackle the segmentation task with weak labels, especially the image-level labels. After training a classification network with the given image-level annotations, class activation maps (CAMs) \cite{zhou2016learning} is leveraged to provide location clues of objects. However, CAMs can only activate the most discriminative object part as segmentation seeds, which are usually sparse and incomplete. Therefore, recent WSSS methods typically concentrate on mining the integral object regions for generating high-quality pixel-level pseudo labels.

 The latest progress of WSSS is mainly driven by multi-stage methods \cite{wei2018revisiting,ahn2018learning,ahn2019weakly,jiang2019integral,li2021pseudo,xu2021leveraging,yao2021non}. Though achieving superior performance, these approaches usually need to train multiple models for different purposes, such as object activation, refinement, and segmentation. The complicated training pipeline will significantly slow down the efficiency. Besides the recent progress of multi-stage methods, several single-stage approaches \cite{papandreou2015weakly,pinheiro2015image,hong2016learning,zhang2020reliability,zhang2021adaptive} have also been explored for efficient training in recent years. For example, RRM \cite{zhang2020reliability} and SSSS \cite{araslanov2020single} train the parallel segmentation branch while leveraging the image classification branch to generate CAMs for obtaining reliable pseudo labels.  While these approaches focus on generating desirable pseudo labels by relying on CAMs, we offer an alternative for pseudo label generation by resorting to the readily accessible saliency maps. Different from existing works that usually leverage saliency maps for providing boundary information in the pseudo label generation step \cite{jiang2019integral,wu2021embedded} or designing a saliency loss in a joint training manner \cite{lee2021railroad}, we directly turn the saliency maps into class-aware pseudo labels for segmentation model training. 
 
 Our work is inspired by STC \cite{wei2016stc} which proposes to learn with the saliency maps of simple images with a single category of major object(s). Different from \cite{wei2016stc} that adopts a three-stage simple to complex framework to learn initial-DCNN, enhanced-DCNN, and powerful-DCNN, we propose to learn the segmentation model in a single-stage scenario from the perspective of alleviating the noisy label and multi-class generalization issues of saliency-generated pseudo labels.
 
 \subsection{Learning with Noisy Labels}
The recent success of deep learning is dependent on the availability of massive and carefully labeled data. In the presence of noisy labels, the performance of deep networks will be significantly impaired due to the strong memorization power of DNNs \cite{arpit2017closer}. In the classification task, loss correction and sample selection are the most popular families for learning with noisy labels. While the former concentrates on designing noise-tolerant loss functions \cite{reed2014training}, the latter aims to select correctly-labeled samples for training by finding proper sample selection criteria \cite{han2018co,wei2020combating}. Since the salient region is not always consistent with the object area (shown in Fig~\ref{fig_moti}), directly training the segmentation model with our saliency-generated pseudo labels will inevitably suffer from the noisy label problem. In our work, we closely examine how the noise is introduced in our WSSS task and divide noise in our saliency-generated pseudo labels into two different granularities. Accordingly,  motivated by sample selection, we propose an online noise filtering module with class adaptive thresholds and a progressive noise detection module to tackle the image-level and pixel-level noise, respectively.

 \subsection{Domain Adaptation}
 
Domain adaptation techniques aim to transfer the knowledge learned from a labeled domain to another unlabeled one through alleviating domain-shift \cite{bousmalis2017unsupervised,hoffman2018cycada,long2015learning,tsai2018learning}. According to which space they are applied, domain adaptation methods can be divided into three categories: input-level, feature-level, and output-level. The input-level approaches typically transfer the style of labeled data to that of unlabeled one via data generation \cite{bousmalis2017unsupervised,hoffman2018cycada}. And adversarial learning prevails in feature- and output-level methods to narrow the domain gap through extracting domain-invariant representations \cite{long2015learning,tsai2018learning}. For our pseudo labels generation, saliency maps can only be well approximated as pseudo labels for simple images that contain one category of objects. In other words, we can only provide labeled simple images and unlabeled complex ones for network training. Motivated by the style alignment through data generation, we propose to generate synthetic complex images from simple ones to alleviate the data distribution gap caused by the difference in category complexity. Besides, we introduce adversarial learning into our WSSS task to enforce the network predictions of complex images to be as close as possible to the pseudo ground truth labels of simple images.

\begin{figure*}[h]
	\centering
	\includegraphics[width=1.0\linewidth]{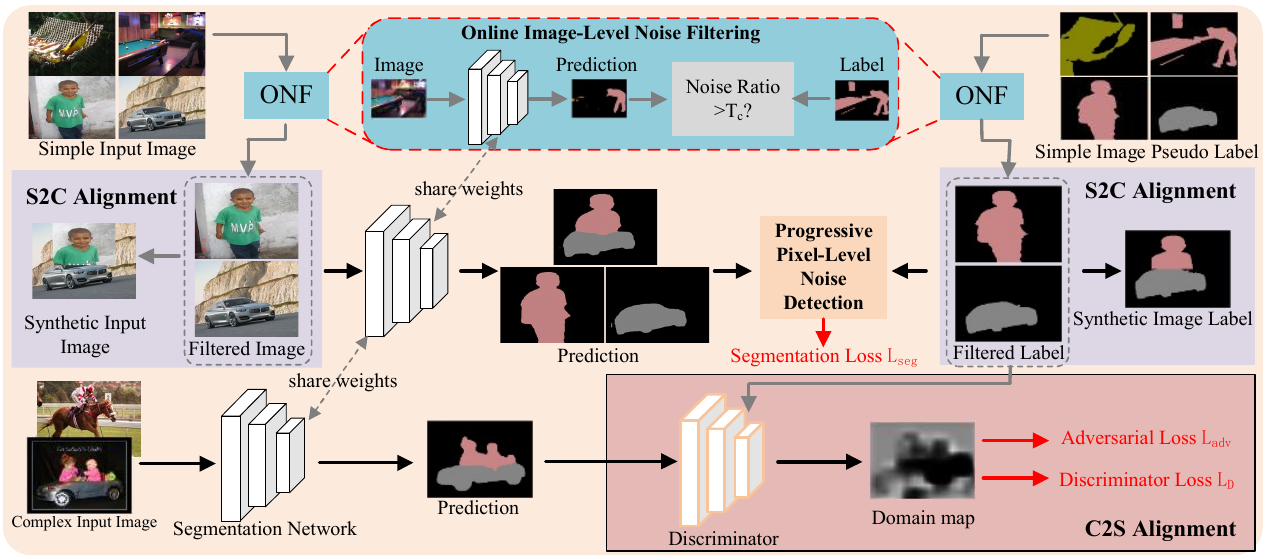}
	\caption{The architecture of our proposed approach. Considering that the saliency-generated pseudo labels are noisy, we propose an online image-level noise filtering module to identify labels with image-level noise and directly discard them during training to avoid misleading gradient back-propagation. Then we propose a progressive pixel-level noise detection module (detailed in Section~\ref{sec_pnd}) to pick out the noisy pixels with a gradually stricter threshold. To narrow the data discrepancies between simple and complex images, a bidirectional alignment mechanism is proposed to reduce the data distribution gap at both input and output space with simple-to-complex (S2C) image synthesis and complex-to-simple (C2S) adversarial learning.}
	\label{fig_framework}
\end{figure*}

\section{The Approach}
\label{approach}

In this section, we propose an end-to-end multi-granularity denoising and bidirectional alignment (MDBA) method with saliency-generated pseudo labels to address the WSSS task. MDBA is illustrated in Fig~\ref{fig_framework}. With the help of image-level labels, we turn class-agnostic saliency maps of simple images into class-aware pseudo labels (shown in Fig~\ref{fig_moti}). Then we directly train a segmentation model with these labeled simple images and unlabeled complex images. Since these saliency-generated pseudo labels for simple images are inevitably noisy, we propose an online image-level noise filtering module and a progressive pixel-level noise detection module to tackle noise of different granularities. To guarantee the generalization of the segmentation model trained with simple images to the complex ones, we propose a bidirectional alignment mechanism to narrow the data distribution gap at both input and output levels with complex image synthesis and adversarial training, respectively.

\subsection{Online Image-Level Noise Filtering}
\label{sec_onf}	
Since saliency maps from the off-the-shelf saliency detection model can offer rich boundaries, they have been widely used in the WSSS task to alleviate the issue of inaccurate object boundaries. Given the class information provided in image-level labels, we turn class-agnostic saliency maps into class-aware pseudo labels and propose to directly train a segmentation model in an elegant end-to-end single-stage manner. However, the detected salient regions might not fit the target objects perfectly or might even include other irrelevant non-target objects, leading to two different granularities of noise in these saliency-generated pseudo labels: 1) image-level noise and 2) pixel-level noise. We notice that labels with image-level noise usually have a high noise rate and tend to deteriorate network training. Therefore, as shown in the upper part of Fig~\ref{fig_framework}, we propose an online image-level noise filtering module, which identifies labels with image-level noise and directly discards them during training to avoid misleading gradient back-propagation. 

After obtaining predictions from the segmentation network, we propose to measure the noise ratio of pseudo labels by calculating the IoU between pseudo labels and network predictions. We define the noise ratio as: $1-$IoU. The lower the IoU, the noisier the pseudo label will be. If the noise ratio is higher than a predefined threshold, we treat the label as the image-level noisy label. We notice that though choosing a common threshold for all the classes is inappropriate, manually searching different thresholds for each category is also infeasible. Therefore, we propose a class-adaptive thresholding technique based on a dataset-level evaluation to determine the threshold for each class. Specifically, after the warm-up training, we first evaluate the network prediction on the whole training set with the saliency-generated pseudo labels to obtain the accuracy of each category. Then we calculate the class-adaptive threshold $T_{c}$ as follows:
\begin{equation}
	T_{c} = 1 -  \left ( a_{c}-\alpha \right ).
	\label{eqa_tc}
\end{equation}
Here, $a_{c}$ denotes the accuracy measure by IoU for class $c$. We adopt the same standard evaluation code to calculate $a_{c}$, where pixels are counted at the dataset level for IoU calculation. Note that since the pseudo labels are noisy, the obtained accuracy $a_{c}$ is higher than the real accuracy. Therefore, as shown in Eq~\eqref{eqa_tc}, for each class $c$, we treat the accuracy $a_{c}$ as a reference and choose a slightly lower value as the real accuracy for the calculation of threshold $T_{c}$ to filter image-level noisy labels.  Specifically, for each simple image, if its noise ratio is higher than $T_{c}$, we treat its pseudo label as the unreliable one and skip its prediction-(pseudo) label pair for loss calculation and gradient back-propagation. We empirically set $\alpha =0.1$.

\subsection{Progressive Pixel-Level Noise Detection}
\label{sec_pnd}

\begin{figure}[h]
	\centering
	\includegraphics[width=1.0\linewidth]{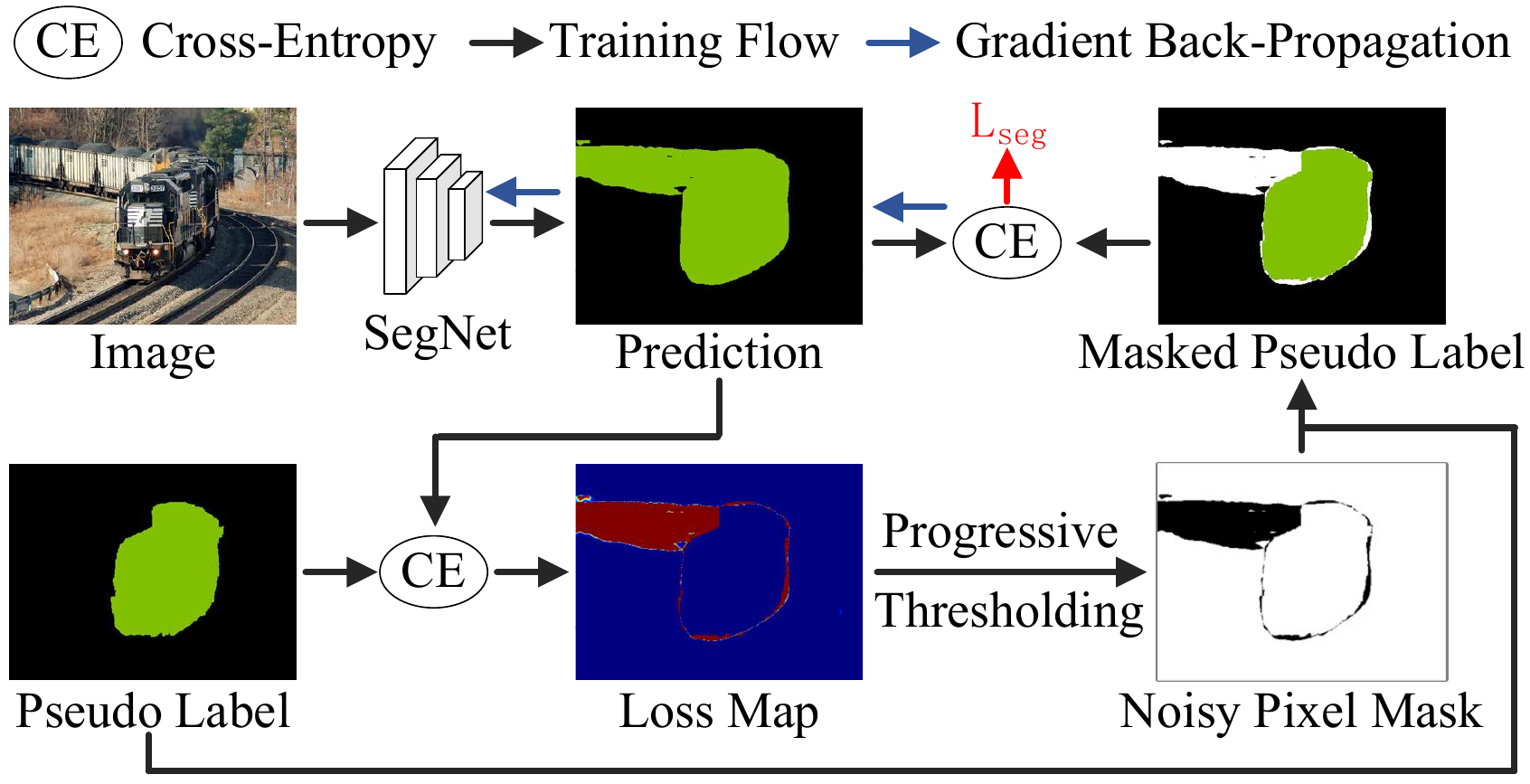}
	\caption{Our proposed progressive pixel-level noise detection module. }
	\label{fig_ppnd}
\end{figure}

After removing image-level noisy labels, we deal with pixel-level noise in this subsection. The deep neural network tends to first learn the simple and general patterns of the real data before fitting the noise \cite{arpit2017closer}. Consequently, at the early training stage, the network predictions will differ significantly with incorrectly labeled labels, resulting in larger cross-entropy losses. Therefore, the loss-based sample selection criterion has been widely adopted in the classification task with noisy labels \cite{han2018co}: instances with higher loss values are more likely to have noisy labels. For our segmentation task with pixel-level noise, as shown in Fig~\ref{fig_ppnd}, we treat each pixel as an instance and pick out noisy pixels according to the values of their corresponding cross-entropy losses:
\begin{equation}
	L_{h,w}= -\sum_{c\in C}Y_{h,w}^{c } \cdot \log P_{h,w}^{c}. 
	\label{eq_pix}
\end{equation} 
$P_{h,w}^{c}$ and $Y_{h,w}^{c }$ are the network prediction and pseudo label for the pixel at location $(h,w)$, respectively; $C$ is the number of classes. After obtaining the cross-entropy loss for each pixel, we propose a progressive pixel-level noise detection module to pick out noisy pixels. Specifically, we obtain a noisy pixel mask $A$ with a predefined threshold $T$:

\begin{equation}
	A_{h,w} = \left\{
	\begin{array}{ll}
		0,  & \text {if}\ L_{h,w}>T \\
		1, & \text { otherwise }
	\end{array}.
	\right.
	\label{eq_mask}
\end{equation}
As shown in Eq~\eqref{eq_mask}, noisy pixels with a high cross-entropy loss are marked as 0 in $A$, and clean pixels are marked as 1. Considering that the network will become more robust as the training goes on, we propose to dynamically adjust the selection threshold according to the current iteration step $t$ as follows:
\begin{equation}
	T = \left\{
	\begin{array}{ll}
		+\infty,  & 1\leq t\leq t_{w} \\
		T_{h} - \left \lfloor \frac{t-t_{w}}{t_{s}} \right \rfloor \cdot \Delta T, &  t_{w}<t\leq t_{max} 
	\end{array}.
	\right.
	\label{eq_thr}
\end{equation}

\begin{equation}
	\Delta T = \frac{T_{h}-T_{l}}{t_{max}-t_{w}} .
	\label{eq_dt}
\end{equation}
Here, $t_{w}$ is the warm-up step and $t_{max}$ is the max step. $t_{s}$ is stride that triggers the decrease of the threshold $T$. $T_{h}$ and $T_{l}$ are the hyper-parameters that denote the highest and lowest value of the threshold $T$.

After obtaining the noisy pixel mask, the final primary segmentation loss is defined as follows:
\begin{equation}
	L_{seg}= \frac{\sum_{h,w}A_{h,w} \cdot L_{h,w}}{\sum_{h,w}A_{h,w}}.
	\label{eq_seg}
\end{equation}

For the training of the segmentation network, we also add a classification constraint on the network output before the softmax layer. Specifically, after removing the background component, we add a global average pooling layer to obtain the standard classification logits for the foreground classes. Then we adopt the multi-label soft margin loss as the classification loss:
\begin{equation}\small
	L_{cls}=-\frac{1}{C^{'}} \sum_{c=1}^{C^{'}} y_{c} \log \sigma\left(p_{c}\right)+\left(1-y_{c}\right) \log \left[1-\sigma\left(p_{c}\right)\right].
\end{equation}
Here, $p_{c}$ is the classification prediction for the $c$-th foreground class. $\sigma\left(\cdot \right)$ is the sigmoid function, and $C^{'}$ is the total number of foreground classes. $y_{c}$ is the image-level label for the $c$-th class. Its value is 1 if the class is present in the image; otherwise, its value is 0. Note that since the image-level labels are ground truth without noise, the classification loss is applied to all training images, including those identified by the online image-level noise filtering module in Section \ref{sec_onf} and discarded for the segmentation task. The added classification constraint helps the network locate the object region, which significantly facilitates the segmentation of training samples, especially for complex images and simple ones with image-level noise that lack the guidance of segmentation loss.

\subsection{Bidirectional Alignment}
Apart from the noise issue, we find that saliency maps can only be well approximated as pseudo labels for simple images that contain one category of objects. For complex images having two or more categories of objects, we cannot determine which category label should be assigned to each pixel of the salient area. However, due to the substantial distribution discrepancies between simple and complex images, learning a segmentation network with only simple images cannot generalize well to complex ones. Therefore, we propose a bidirectional alignment mechanism to alleviate the multi-class generalization problem by transferring knowledge from the labeled simple images to unlabeled complex ones. 

{\textbf{Simple-to-Complex Input Alignment.}} Distribution alignment in raw pixel space has been widely used in both classification and segmentation tasks to reduce domain gap \cite{bousmalis2017unsupervised,hoffman2018cycada}. However, these methods mainly focus on translating source data to the style of a target domain for reducing the low-level appearance variance, which is not applicable to alleviate our distribution gap caused by the difference of category complexity. We notice that style transfer is actually a kind of data augmentation to construct training data that seem like the data from target domain. From this point of view, as shown in Fig~\ref{fig_framework}, we propose to generate synthetic complex images from the simple ones via the CutMix \cite{yun2019cutmix} augmentation strategy to approximate the data distribution of complex images at the input level. The goal of CutMix is to generate a new pair of training image and label $\left ( \tilde{x},\tilde{y} \right )$ by combining two samples $\left ( x_{A},y_{A} \right )$ and $\left ( x_{B},y_{B} \right )$. Extending the CutMix from classification to segmentation task, the combining operation can be defined as follows:
\begin{equation}
	\begin{array}{l}
		\tilde{x} = M \odot x_{A} + \left (1- M \right ) \odot x_{B} \\
		\tilde{y} = M \odot y_{A} + \left (1- M \right ) \odot y_{B}
	\end{array}.
	\label{eq_cutmix}
\end{equation}
Here, $M \in \left \{ 0,1 \right \}^{H\times W}$ is a binary mask indicating where to drop out and fill in for two images (labels), and $\odot$ is element-wise multiplication.

{\textbf{Complex-to-Simple Output Alignment.}}
After aligning the data distribution in input space, to help the network learned with labeled images generalize well to complex unlabeled data, we utilize the adversarial training in the output space \cite{tsai2018learning} to further mitigate the distribution gap between simple and complex images. As shown in the lower part of Fig~\ref{fig_framework}, we introduce adversarial learning to enforce the network predictions of complex images to be as close as possible to the pseudo ground truth labels of simple images. To facilitate the convergence of the network training, predictions of simple images are also involved in adversarial learning. Given the predicted probability map 
$P$ and pseudo ground truth label of the simple image $Y_{s}$ with size $H\times W\times C$, where $C$ is the number of classes, we train a fully convolutional discriminator $D$ with the discriminative loss as follows:

\begin{equation}\small
	{L}_{D}=-\sum_{h, w}\left(1-z\right) \log \left(1-D\left(P\right)^{(h, w)}\right)
	+ z \log \left(D\left({Y}_{s}\right)^{(h, w)}\right),
	\label{eq_dis}
\end{equation}
where $z$ = 0 if the sample is drawn from the predicted probability maps, and $z$ = 1 for the sample from the pseudo ground truth of simple images. We adopt the one-hot encoding scheme to convert the pseudo ground truth with discrete labels to a $C$-channel probability map.

To make the distribution of network prediction closer to that of pseudo ground truth $Y_{s}$, we train the segmentation network with an adversarial loss as follows:

\begin{equation}
	{L}_{adv}=- \sum_{h, w} \log \left(D\left(P\right)^{(h, w)}\right).
	\label{eq_adv}
\end{equation}
This loss is designed to train the segmentation model and fool the discriminator by maximizing the probability of network predictions being considered as pseudo ground truth labels of simple images.

\subsection{Training Objective}
The overall loss function for training our network can be written as:
\begin{equation}
	L\left ( G,D \right ) = L_{seg} + L_{cls} + L_{D} + \lambda_{adv}{L}_{adv}.
	\label{eq_all}
\end{equation}
Here, $\lambda_{adv}$ is the hyper-parameter that controls the relative importance of the adversarial loss. The training objective of our approach is:
\begin{equation}
	G^{*}, D^{*}= \arg\ \min \limits_{G}\, \max \limits_{D}\, L\left ( G,D \right ).
	\label{eq_obj}
\end{equation}
We solve Eq~\eqref{eq_obj} by alternately optimizing the segmentation network $G$ and the discriminator $D$.

\section{Experiments}
\label{experiments}

\subsection{Implementation Details}
For single-stage training, we adopt the DeepLab-v2 \cite{chen2017deeplab} framework with ResNet-101 \cite{he2016deep} model pre-trained on the ImageNet dataset \cite{deng2009imagenet} and MS COCO \cite{lin2014microsoft} as our segmentation network. The COCO-pretrained model is directly converted from the Caffe model released by the DeepLab-v2 authors \cite{papandreou2015weakly,chen2017deeplab}. The discriminator network consists of five convolutional layers with kernel size 4 $\times$ 4, where the channel number is $\left \{64, 128, 256, 512, 1\right \}$, respectively. Except for the last functional layer, each convolutional layer is followed by a Leaky-ReLU parameterized by 0.2. 

We use SGD and ADAM as the optimizer for the segmentation and discriminator network, respectively. The momentum and weight decay of SGD are 0.9 and $10^{-4}$. The initial learning rate is set to $ 2.5 \times 10^{-4}$ and is decreased using the polynomial decay with a power of 0.9. For the ADAM optimizer, the same polynomial decay is used while the initial learning rate is $10^{-4}$. We set batch size = 10. For progressive pixel-level noise detection module, we empirically set $t_{w} =1000$, $t_{max} = 11000$, $t_{s}=1000$, $T_{h} = 1.2$ and $T_{l}=0.8$. For adversarial loss in Eq~\eqref{eq_all}, we empirically set $\lambda_{adv}=0.001$. The saliency maps used in this paper are provided by OAA \cite{jiang2019integral}, which are generated by a pretrained saliency detection model \cite{hou2017deeply}.

For the two-step training, we leverage the segmentation network obtained by our proposed method to generate pseudo labels for retraining the network. For fair comparison, we use the segmentation code released by NSROM \cite{yao2021non} and EPS \cite{lee2021railroad} to evaluate the quality of pseudo labels generated by our proposed method. Results with both VGG and ResNet backbones are reported. 

\subsection{Datasets and Evaluation Metrics}
Following previous works, we evaluate our approach on the PASCAL VOC 2012 dataset \cite{everingham2010pascal} and COCO dataset \cite{lin2014microsoft}. As the most popular benchmark for WSSS, the PASCAL VOC 2012 dataset contains 21 classes (20 object categories and the background) for semantic segmentation. There are 10,582 training images, which are expanded by \cite{hariharan2011semantic}, 1,449 validation images, and 1,456 test images. COCO dataset is a more challenging benchmark with 80 semantic classes and the background. Following previous works \cite{wang2020weakly,Li2021GroupWiseSM,zhang2020causal}, we use the default train/val splits (80k images for training and 40k for validation) in the experiment. Standard mean intersection over union (mIoU) is taken as the evaluation metric for semantic segmentation.

\begin{table}[t]
	
	\setlength{\tabcolsep}{4mm}
	\renewcommand\arraystretch{1.2}
	\centering
	\caption{Quantitative comparisons to previous state-of-the-art \textbf{single-stage} approaches on PASCAL VOC 2012 dataset. * denotes model is pre-trained on MS-COCO. }
	\begin{tabular}{{l}*{3}{c}}
		\hline
		Methods & Publication &  Val & Test\\
		\hline
		\textbf{Single-Stage}\\
		
		EM \cite{papandreou2015weakly} &ICCV15&38.2 &39.6\\
		MIL-LSE \cite{pinheiro2015image} &CVPR15&42.0 &40.6\\ 
		CRF-RNN \cite{roy2017combining} &CVPR17&52.8 &53.7\\ 
		TransferNet \cite{hong2016learning} &CVPR16&52.1 &51.2\\
		WebCrawl \cite{hong2017weakly} &CVPR17&58.1 &58.7\\
		RRM \cite{zhang2020reliability} &AAAI20&62.6 &62.9\\ 
		SSSS \cite{araslanov2020single} &CVPR20&62.7 &64.3\\
		Zhang \etal~\cite{zhang2021adaptive}&ACMMM21&63.9 &64.8\\   
		Chen \etal~\cite{chen2021end}&ACMMM21&63.6&65.7\\
		\textbf{MDBA (Ours)} &-&\textbf{66.3} &\textbf{66.4}\\	
		\textbf{MDBA (Ours)}*  &-&\textbf{69.5} &\textbf{70.2}\\		
		\hline
	\end{tabular}
	
	\label{tab_st}	
\end{table}

\begin{table}[t]
	\setlength{\tabcolsep}{3.4mm}
	\renewcommand\arraystretch{1.2}
	\centering
	\caption{Parameter comparison to previous state-of-the-art \textbf{single-stage} approaches. }
	\begin{tabular}{{l}*{3}{c}}
		\hline
		 Methods& Publication & Parameter& GPU\\
		\hline
		\textbf{Single-Stage}\\
		
		RRM \cite{zhang2020reliability} & AAAI20& 124M& 44G\\
	SSSS \cite{araslanov2020single}  & CVPR20& 138M& 24G\\
Zhang \etal~\cite{zhang2021adaptive}& ACMMM21& -& 44G\\
Chen \etal~\cite{chen2021end}& ACMMM21& -& 72G\\
\textbf{MDBA (Ours)}& -& \textbf{47M}& \textbf{24G}\\
		\hline
	\end{tabular}
	
	\label{tab_gpu}	
\end{table}

\begin{table*}[t]
	\begin{minipage}{0.48\linewidth}
		\centering
		\setlength{\tabcolsep}{4mm}
		\renewcommand\arraystretch{1.24}
		\caption{Quantitative comparisons to previous state-of-the-art \textbf{multi-stage} approaches on PASCAL VOC 2012 dataset with VGG backbone.}
		\begin{tabular}{{l}*{3}{c}}
			\hline
			Methods & Publication & Val & Test\\
			\hline
			\textbf{Multi-Stage}\\
			STC \cite{wei2016stc}&TPAMI17&49.8 &51.2\\ 
			Hong \etal \cite{hong2017weakly} &CVPR17&58.1 &58.7\\ 
			TPL \cite{kim2017two} &ICCV17&53.1 &53.8\\ 
			GAIN \cite{li2018tell} &CVPR18&55.3 &56.8\\ 
			DSRG \cite{huang2018weakly} &CVPR18&59.0 &60.4\\
			MCOF \cite{wang2018weakly} &CVPR18&56.2 &57.6\\
			AffinityNet \cite{ahn2018learning} &CVPR18&58.4 &60.5\\  
			RDC \cite{wei2018revisiting} &CVPR18&60.4 &60.8\\
			SeeNet \cite{hou2018self} &NIPS18&63.1 &62.8\\
			OAA \cite{jiang2019integral}&ICCV19&63.1 &62.8\\  
			ICD \cite{fan2020learning} &CVPR20&64.0 &63.9\\
			BES \cite{chen2020weakly}&ECCV20&60.1 &61.1\\ 
			Fan \etal \cite{fan2020employing} &ECCV20&64.6 &64.2\\
			Zhang \etal \cite{zhang2020splitting} &ECCV20&63.7 &64.5\\  
			MCIS \cite{sun2020mining}&ECCV20&63.5 &63.6\\ 
			NSROM \cite{yao2021non} &CVPR21&65.5 &65.3\\
			EPS \cite{lee2021railroad}&CVPR21&67.0 &67.3\\
			ECS-Net \cite{sun2021ecs} &ICCV21&62.1 &63.4\\
			L2G \cite{jiang2022l2g}&CVPR22&68.5 &68.9\\
			\hline
			\textbf{Single-Stage}\\
			RRM \cite{zhang2020reliability} (two-step)&AAAI20&60.7 &61.0\\ 
			\textbf{MDBA (Ours)} (two-step)&-&\textbf{68.6} &\textbf{68.5}\\	
			\hline		
		\end{tabular}
		
		\label{tab_vgg_2s}	
	\end{minipage}
	\hfill
	\begin{minipage}{0.48\linewidth}
		\centering
		\setlength{\tabcolsep}{4mm}
		\renewcommand\arraystretch{1.1}
		\caption{\small{Quantitative comparisons to previous state-of-the-art \textbf{multi-stage} approaches on PASCAL VOC 2012 dataset with ResNet backbone. * denotes model is pre-trained on MS-COCO. }}
		\begin{tabular}{{l}*{3}{c}}
			\hline
			Methods & Publication  & Val & Test\\
			\hline
			\textbf{Multi-Stage}\\
			
			
			SEAM \cite{wang2020self} &CVPR20 &64.5 &65.7\\
			ICD \cite{fan2020learning} &CVPR20 &67.8 &68.0\\
			MCIS \cite{sun2020mining} &ECCV20&66.2 &66.9\\
			BES \cite{chen2020weakly}&ECCV20 &65.7  &66.6\\  
			CONTA \cite{zhang2020causal}&NIPS20 &66.1 &66.7\\
			GWSM \cite{Li2021GroupWiseSM} &AAAI21 &68.2&68.5\\
			NSROM \cite{yao2021non} &CVPR21 &68.3 &68.5\\
			OC-CSE \cite{kweon2021unlocking} &ICCV21 &68.4 &68.2\\
			PMM \cite{li2021pseudo} &ICCV21 &68.5 &69.0\\
			AuxSegNet \cite{xu2021leveraging} &ICCV21 &69.0 &68.6\\
			
			\hline
			\textbf{Single-Stage}\\
			RRM \cite{zhang2020reliability} (two-step)&AAAI20 &66.3 &66.5\\ 
			SSSS\cite{araslanov2020single} (two-step)&CVPR20 &65.7 &66.6\\
			
			\textbf{MDBA (Ours)} (two-step)&-&\textbf{70.0} &\textbf{70.2}\\	
			\hline
			\hline
			\textbf{Multi-Stage}\\
			*NSROM \cite{yao2021non} &CVPR21&70.4 &70.2\\
			*EPS \cite{lee2021railroad}&CVPR21 &70.9 &70.8\\
			*EDAM \cite{wu2021embedded}&CVPR21 &70.9 &70.6\\
			*CLIMS \cite{xie2022clims}&CVPR22 &70.4 &70.0\\
			*AMN \cite{lee2022threshold}&CVPR22 &70.7 &70.6\\
			*L2G \cite{jiang2022l2g}&CVPR22&72.1 &71.7\\
			*OPC \cite{feng2023weakly} &PRL23 &70.5 &71.8\\
			\hline
			\textbf{Single-Stage}\\
			\textbf{*MDBA (Ours)} (two-step)  &- &\textbf{72.0} &\textbf{71.5}\\		
			\hline	
		\end{tabular}
		
		\label{tab_ms}	
	\end{minipage}
\end{table*}

\subsection{Comparisons to the State-of-the-Arts}
{\textbf{Baselines.}}
In this part, we compare our proposed method with the following state-of-the-art approaches that leverage image-level labels for WSSS: 

\noindent {\textbf{Single-Stage Methods:}} EM \cite{papandreou2015weakly}, MIL-LSE \cite{pinheiro2015image}, CRF-RNN \cite{roy2017combining}, TransferNet \cite{hong2016learning}, WebCrawl \cite{hong2017weakly}, RRM \cite{zhang2020reliability}, SSSS\cite{araslanov2020single}, Zhang \etal~\cite{zhang2021adaptive}, Chen \etal~\cite{chen2021end}.

\noindent {\textbf{Multi-Stage Methods:}} BFBP \cite{saleh2016built}, SEC \cite{kolesnikov2016seed}, DSRG \cite{huang2018weakly}, OAA \cite{jiang2019integral}, IAL \cite{wang2020weakly}, IRN \cite{ahn2019weakly}, SEAM \cite{wang2020self}, ICD \cite{fan2020learning}, BES \cite{chen2020weakly}, Fan \etal \cite{fan2020employing}, Zhang \etal~ \cite{zhang2020splitting}, MCIS \cite{sun2020mining},  CONTA \cite{zhang2020causal}, GWSM \cite{Li2021GroupWiseSM}, NSROM \cite{yao2021non}, EPS \cite{lee2021railroad}, EDAM \cite{wu2021embedded}, OC-CSE \cite{kweon2021unlocking}, PMM \cite{li2021pseudo}, AuxSegNet \cite{xu2021leveraging}, L2G \cite{jiang2022l2g}, OPC \cite{feng2023weakly}.

{\textbf{Experimental Results on PASCAL VOC 2012 Dataset.}} In Table~\ref{tab_st}, we provide performance comparisons with other single-stage methods for WSSS on PASCAL VOC 2012 dataset. As can be seen, our approach achieves better results than other state-of-the-art methods. Specifically, our segmentation results reach the mIoU of 66.3\% and 66.4\% on the validation and test sets, respectively. Our method outperforms the recent approaches of RRM \cite{zhang2020reliability}, SSSS\cite{araslanov2020single}, Zhang \etal~\cite{zhang2021adaptive} and Chen \etal~\cite{chen2021end} by more than 2.4\% on the PASCAL VOC validation set and 0.7\% on the test set. When training the segmentation network with the COCO pre-trained weights, we can further reach the mIoU of 69.5\% and 70.2\% on the validation and test set, respectively.

We also compare our proposed method with recent single-stage approaches from the perspectives of model size and the GPU environment. Since the existing single-stage methods usually need to simultaneously maintain several networks for classification, CAMs refinement and segmentation, they tend to involve more trainable parameters. As shown in Table~\ref{tab_gpu}, the networks adopted in RRM \cite{zhang2020reliability} and SSSS \cite{araslanov2020single} contain more than 124M parameters. In contrast, the size of our model is only 47M, much smaller than previous single-stage approaches. For GPU consumption, the implementation of RRM \cite{zhang2020reliability} and the work of Zhang \etal~\cite{zhang2021adaptive} requires a 4-GPU environment of 44G GPU memory. Training the method proposed by Chen \etal~\cite{chen2021end} even needs a platform with 72G GPU memory. Such a high GPU requirement significantly raises the threshold of the single-stage method. By contrast, our approach can achieve substantial improvement with an easily accessible GPU environment (actual GPU consumption of about 18G). Therefore, adopting a non-CAMs-based approach with saliency-generated pseudo labels might be a more economical option for single-stage methods.

Following the work of RRM \cite{zhang2020reliability} and SSSS\cite{araslanov2020single}, we also extend our method to a two-step (two-stage) framework to show the effectiveness and scalability of our approach in Table~\ref{tab_vgg_2s} and \ref{tab_ms}. Though the work of SSSS\cite{araslanov2020single} uses the framework of DeepLabv3+ \cite{chen2018encoder}, we still adopt the DeepLab-v2 \cite{chen2017deeplab} framework as our segmentation network for fair comparisons with RRM \cite{zhang2020reliability} and other multi-stage approaches. Note that denoising and alignmnet are not adopted in the second step training. As shown in Table~\ref{tab_vgg_2s}, leveraging our method to generate more accurate pseudo labels to train a segmentation network with VGG backbone, we can reach 68.6\% and 68.5\% on PASCAL VOC validation and test sets, respectively. Our results outperform RRM \cite{zhang2020reliability} by more than 7.5\%. As shown in Table~\ref{tab_ms}, with the ResNet backbone, we can reach 70.0\% and 70.2\% on PASCAL VOC validation and test sets, respectively. Our approach can outperform RRM \cite{zhang2020reliability} and SSSS\cite{araslanov2020single} by more than 3.6\% mIoU.

\begin{figure*}[t]
	\centering
	\includegraphics[width=1.0\linewidth]{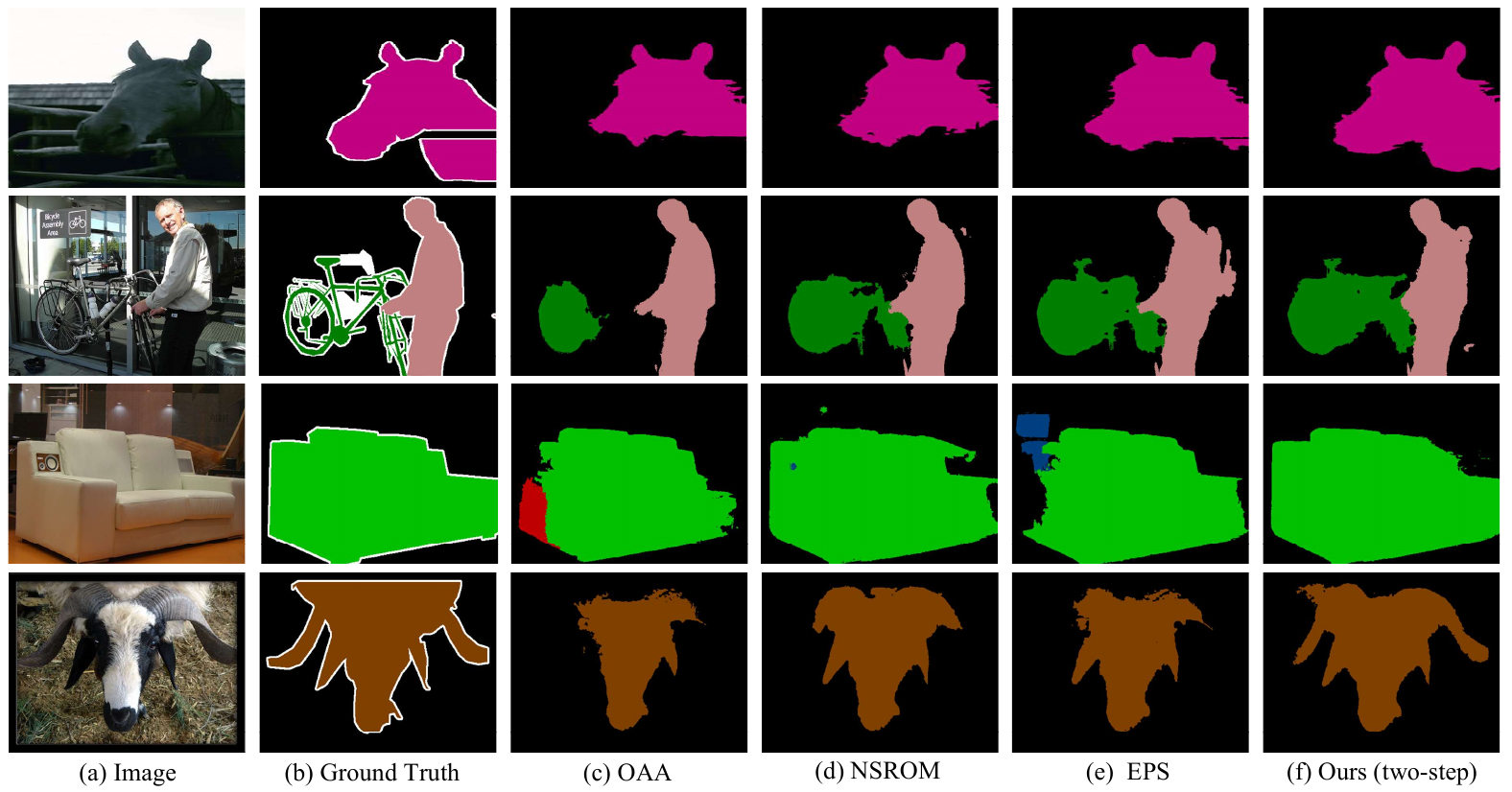}
	\caption{Comparisons of our two-step training results with other state-of-the-art multi-stage approaches on PASCAL VOC 2012 validation set. For each (a) image, we show (b) ground truth, and the results of (c) OAA \cite{jiang2019integral}. (d) NSROM \cite{yao2021non}, (e) EPS \cite{lee2021railroad}, (f) our two-step training. Best viewed in color.}
	\label{fig_multi_comp}
\end{figure*}

Besides comparisons with single-stage methods, we also compare our two-step results with state-of-the-art multi-stage approaches. As shown in Table~\ref{tab_vgg_2s}, with the VGG backbone, our method achieves competitive performance compared to L2G \cite{jiang2022l2g} and significantly outperforms NSROM \cite{yao2021non} and EPS \cite{lee2021railroad} by more than 1.2\% mIoU. Leveraging the ResNet backbone, as shown in Table~\ref{tab_ms}, our method can outperform the recent work of PMM \cite{li2021pseudo} and AuxSegNet \cite{xu2021leveraging} by more than 1.0\%, even though the two approaches iteratively train the segmentation network for multiple rounds. When leveraging the COCO pre-trained weights for the network initialization, our two-step performance can further reach the mIoU of 72.0\% on the PASCAL VOC validation set and 71.5\% on the test set. As we can see, while achieving competitive results to the state-of-the-art multi-stage approaches like L2G \cite{jiang2022l2g} and OPC \cite{feng2023weakly}, our MDBA can outperform EPS \cite{lee2021railroad} and EDAM \cite{wu2021embedded} by about 1.0\%. Our results illustrated in Table~\ref{tab_vgg_2s} and Table~\ref{tab_ms} demonstrate the effectiveness of our proposed approach. Fig.~\ref{fig_multi_comp} presents the comparisons of our two-step training results with other state-of-the-art multi-stage approaches on the PASCAL VOC 2012 validation set.

{\textbf{Experimental Results on COCO Dataset.}} In Table~\ref{tab_coco}, we provide performance comparisons with state-of-the-art methods for WSSS on the COCO dataset. Since recent single-stage methods do not report performance on COCO dataset, we compare our method with multi-stage approaches in this part. The results of SEAM \cite{wang2020self} and IRN \cite{ahn2019weakly} are from \cite{zhang2020causal}. As shown in Table~\ref{tab_coco}, when training the segmentation network in an end-to-end manner, our approach can obtain the mIoU of 36.0\%, which is competitive to the results obtained by multi-stage approaches of PMM \cite{li2021pseudo} and OC-CSE \cite{kweon2021unlocking}. While extending our method to a two-step framework, we can obtain the performance of 37.8\%, outperforming the recent work of OC-CSE \cite{kweon2021unlocking} by 1.4\%.

\begin{table}[t]	
	
	\setlength{\tabcolsep}{3mm}
	\renewcommand\arraystretch{1.2}
	\caption{\small{Quantitative comparisons to previous state-of-the-art \textbf{multi-stage} approaches on MS COCO dataset. }}
	\centering
	\begin{tabular}{{l}*{3}{c}}
		\hline
		Methods & Publication  & Val \\
		\hline
		\textbf{Multi-Stage}\\
		BFBP \cite{saleh2016built}  &ECCV16&20.4\\ 
		SEC \cite{kolesnikov2016seed} &ECCV16&22.4\\ 
		DSRG \cite{huang2018weakly} &CVPR18&26.0\\ 
		IRN \cite{ahn2019weakly} &CVPR19 &32.6\\
		IAL \cite{wang2020weakly} &IJCV20 &27.7\\ 
		SEAM \cite{wang2020self} &CVPR20 &31.9\\
		CONTA \cite{zhang2020causal}&NIPS20 &33.4\\
		GWSM \cite{Li2021GroupWiseSM} &AAAI21 &28.4\\
		EPS \cite{lee2021railroad}&CVPR21 &35.7\\
		AuxSegNet \cite{xu2021leveraging} &ICCV21 &33.9\\
		PMM \cite{li2021pseudo} &ICCV21 &35.7\\
		OC-CSE \cite{kweon2021unlocking} &ICCV21 &36.4\\
		\hline
		\textbf{Single-Stage}\\
		\textbf{MDBA (Ours) } &-&\textbf{36.0} \\	
		\textbf{MDBA (Ours) } (two-step) &- &\textbf{37.8} \\	
		\hline
	\end{tabular}
	
	\label{tab_coco}	
\end{table}

\subsection{Complementarity to CAMs-Based Approaches.}

As a non-CAMs-based approach, our MDBA is complementary to those CAMs-based methods. However, due to the GPU limitation, we cannot implement a combined method of ours and other CAMs-based methods. As mentioned above, besides the superiority compared with single-stage methods, our approach can also be easily extended for two-step training. Therefore, we resort to the second stage training to demonstrate the complementarity of our method. Specifically, we leverage the pseudo labels generated by ours and latest CAMs-based approaches (\eg, EPS \cite{lee2021railroad} and L2G \cite{jiang2022l2g}) to train a segmentation network with loss as follows:
\begin{equation}
\mathcal{L} = \lambda_{MDBA} \mathcal{L}_{seg}^{MDBA} +  \lambda_{CAM} \mathcal{L}_{seg}^{CAM}.
\label{eq_joint}
\end{equation}
Here, $\lambda_{MDBA}$ and  $\lambda_{CAM}$ are the hyper-parameters that control the relative importance of pseudo labels generated by ours and latest CAMs-based approaches \cite{lee2021railroad,jiang2022l2g}. We simply set $\lambda_{MDBA}=\lambda_{CAM}=0.5$ for equal importance. As shown in Table~\ref{tab_comp}, by combining the pseudo labels of both CAMs and non-CAMs-based approaches, we can obtain 73.5\% mIoU on the PASCAL VOC validation set, which outperforms L2G \cite{jiang2022l2g} and our MDBA by 1.4\% and 1.5\%, respectively. Besides, the combined method can arrive at the mIoU of 73.1\% on the test set. The consistent and substantial improvement compared to the performance of CAMs-based methods \cite{lee2021railroad,jiang2022l2g} and our MDBA verifies the superiority of making good use of the complementarity of both CAMs and non-CAMs-based approaches.

\subsection{Ablation Studies}

\begin{figure*}[ht]
	\centering
	\includegraphics[width=1.0\linewidth]{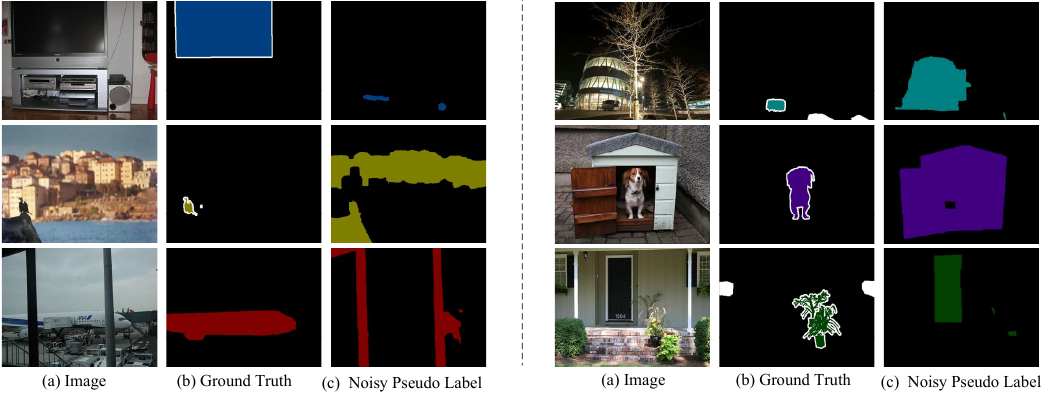}
	\caption{Example of image-level noise filtered by our online noise filtering module on PASCAL VOC 2012 training set. For each (a) image, we show (b) ground truth, and (c) saliency-generated pseudo label (image-level noise). Best viewed in color.}
	\label{fig_flt}
\end{figure*}

\begin{table*}[t]
	\begin{minipage}{0.48\linewidth}
		\setlength{\tabcolsep}{0.5mm}
		\renewcommand\arraystretch{1.22}
		\centering
		\caption{Results of combining CAMs-based and non-CAMs-based methods on PASCAL VOC 2012 dataset. * denotes model is pre-trained on MS-COCO.}
		\begin{tabular}{{l}*{3}{c}}
			\hline
			Methods & Publication & Val & Test\\
			\hline
			\textbf{CAMs-Based}\\
			*EPS \cite{lee2021railroad}&CVPR21&70.9 &70.8\\
			*L2G \cite{jiang2022l2g}&CVPR22&72.1 &71.7\\
			\hline
			\textbf{Non-CAMs-Based}\\
			\textbf{*MDBA (Ours)} (two-step) &-&\textbf{72.0} &\textbf{71.5}\\		
			\hline
			\textbf{Combined}\\
			\textbf{*MDBA (Ours) + EPS} (two-step) &-&\textbf{72.8} &\textbf{72.9}\\
			\textbf{*MDBA (Ours) + L2G} (two-step) &-&\textbf{73.5} &\textbf{73.1}\\	
			\hline
		\end{tabular}
		
		\label{tab_comp}	
	\end{minipage}
	\hfill
	\begin{minipage}{0.48\linewidth}
		\centering
		\renewcommand\arraystretch{1.3}
		\setlength{\tabcolsep}{3mm}
		\caption{Element-wise component analysis. ONF: online noise filtering, PND: progressive noise detection, S2C: simple-to-complex input alignment, C2S: complex-to-simple output alignment, CRF: CRF post-processing. }
		
		\begin{tabular}{*{7}{c}}
			\hline
			Base&ONF&PND&S2C&C2S&CRF&mIoU\\
			\hline
			\checkmark&&&&&&61.2\\
			\checkmark&\checkmark&&&&&62.8\\
			\checkmark&\checkmark&\checkmark&&&&65.4\\
			\checkmark&\checkmark&\checkmark&\checkmark&&&67.9\\	
			\checkmark&\checkmark&\checkmark&\checkmark&\checkmark&&69.0\\
			\checkmark&\checkmark&\checkmark&\checkmark&\checkmark&\checkmark&69.5\\	
			\hline
		\end{tabular}
		
		\label{tab_element}
	\end{minipage}
\end{table*}

\textbf{Element-Wise Component Analysis.}
In this part, we demonstrate the contribution of each component proposed in our approach for end-to-end WSSS. The experimental results on the validation set of PASCAL VOC are given in Table~\ref{tab_element}. As we can see, by leveraging our proposed online noise filtering module to remove the image-level noisy labels during training, we can improve the segmentation result from 61.2\% to 62.8\%. Fig.~\ref{fig_flt} shows some example of the image-level noise that is filtered by our online noise filtering module on PASCAL VOC 2012 training set. With our proposed progressive pixel-level noise detection module to pick out noisy pixels, we obtain another 2.6\% performance gain. By generating synthetic complex images from the simple ones for simple-to-complex alignment at the input level, the mIoU arrives at 67.9\%. Applying adversarial learning to encourage the model to predict segmentation outputs of unlabeled complex images close to the pseudo ground truth distributions, complex-to-simple output alignment contributes to another 1.1\% mIoU performance gain and improves the segmentation result to 69.0\%. Post-processing our model by dense CRF \cite{krahenbuhl2011efficient} finally yields the performance of 69.5\%.

\begin{figure*}[t]
	\centering
	\includegraphics[width=\linewidth]{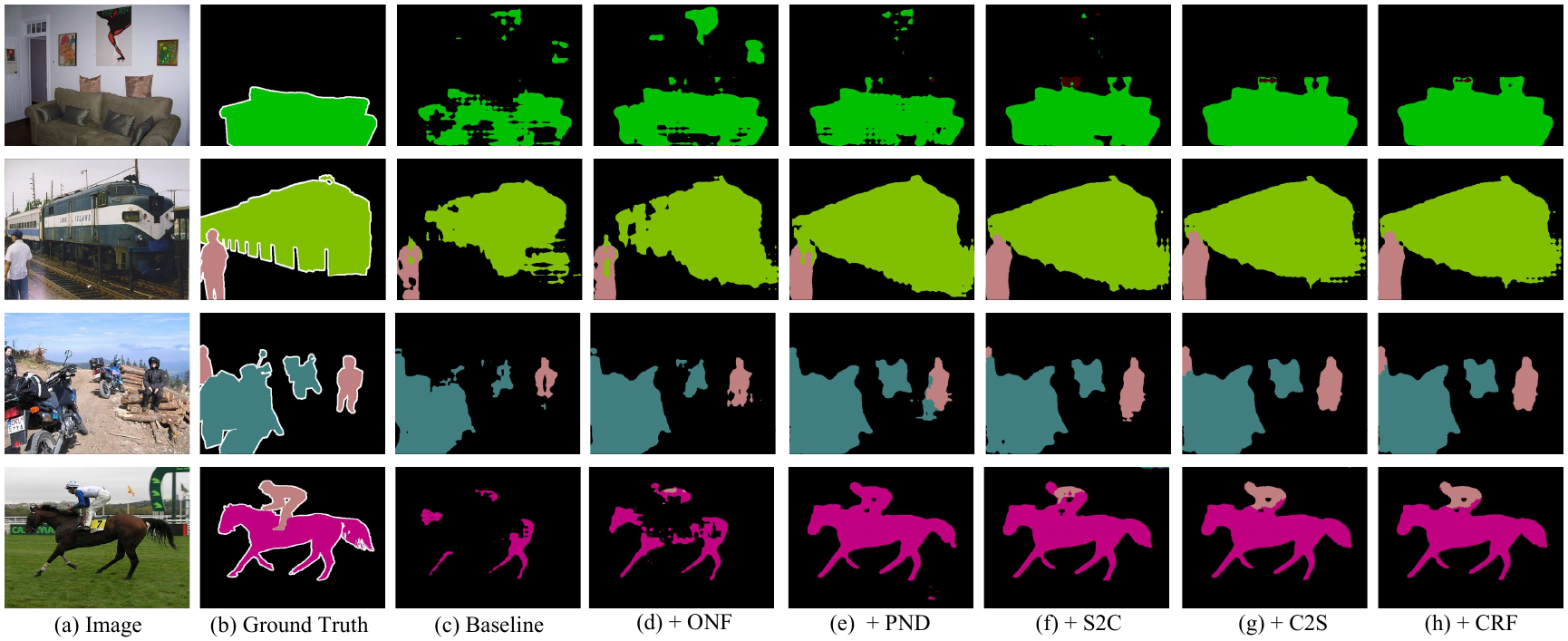}
	\caption{Example segmentation results on PASCAL VOC 2012 validation set. For each (a) image, we show (b) GT (ground truth), results of (c) baseline, (d) + ONF (online image-level noise filtering), (e) + PND (progressive pixel-level noise detection), (f) + S2C (simple-to-complex alignment via complex image synthesis), (g) + C2S (complex-to-simple output alignment via adversarial learning), and (h) + CRF.}
	\label{fig_result}
\end{figure*}

Some qualitative segmentation examples on the PASCAL VOC 2012 validation set can be viewed in Fig~\ref{fig_result}. As can be seen, with our proposed online noise filtering module (ONF), the network can predict much more precise boundaries for the objects (\eg the sofa and train in the first and second rows). Leveraging our proposed progressive noise detection module (PND) to remove noisy pixels, we can get tidier segmentation maps further. The last two rows show that learning a segmentation network with only simple images cannot generalize well to complex ones. By aligning the data distribution at the input level with complex image synthesis (S2C), our method can detect the pixels belonging to the person category that are missed Fig~\ref{fig_result} (e). By aligning segmentation predictions to pseudo ground truth distributions of simple images with adversarial learning (C2S), as shown in Fig~\ref{fig_result} (g), we can further narrow the gap between simple and complex images and render the network generalize better to complex images.

\begin{table*}[t]
	\begin{minipage}{0.32\linewidth}
		\centering
		\setlength{\tabcolsep}{6mm}
		\renewcommand\arraystretch{1.0}
		\caption{Parameter sensitivity of $\alpha$ for the online image-level noise filtering module. Results are reported without CRF.}
		
		\begin{tabular}{@{}lc@{}}
			\hline
			$\alpha$  & mIoU \\
			\hline
			0 & 68.7 \\
			0.05 & 68.9 \\
			0.1 & \textbf{69.0} \\
			0.15 & 68.7 \\
			0.2 & 68.5 \\
			0.25 & 68.1 \\
			\hline
		\end{tabular}
		
		\label{tab_alpha}
	\end{minipage}
	\hfill
	\begin{minipage}{0.32\linewidth}
		\centering
		\setlength{\tabcolsep}{6mm}
		\renewcommand\arraystretch{1.0}
		\caption{Parameter sensitivity of $T_{h}$ and $T_{l}$ in progressive noise detection module. Results are reported without CRF.}
		
		\begin{tabular}{@{}ccc@{}}
			\hline
			$T_{h}$ & $T_{l}$ & mIoU \\
			\hline
			1.0 & 0.8 & 68.7 \\
			1.2 & 0.8 & \textbf{69.0} \\
			1.4 & 0.8 & 68.9 \\
			1.6 & 0.8 & 68.5 \\
			1.2 & 0.6 & 68.3 \\
			1.2 & 1.0 & 68.3 \\
			\hline
		\end{tabular}
		
		\label{tab_ps}
	\end{minipage}
	\hfill
	\begin{minipage}{0.32\linewidth}
		\centering
		\setlength{\tabcolsep}{6mm}
		\renewcommand\arraystretch{1.2}
		\caption{Parameter sensitivity of $\lambda_{adv}$ for the adversarial loss. Results are reported without CRF.}
		
		\begin{tabular}{@{}lc@{}}
			\hline
			$\lambda_{adv}$  & mIoU \\
			\hline
			0.01 & 67.9 \\
			0.005 & 68.8 \\
			0.001 & \textbf{69.0} \\
			0.0005 & 68.7 \\
			0.0001 & 68.5 \\
			\hline
		\end{tabular}
		
		\label{tab_adv}
	\end{minipage}
\end{table*}

\textbf{Parameter Analysis.} 
For online image-level noise filtering module, we conduct experiments to study the effect of hyper-parameter $\alpha$ that controls the class-adaptive threshold. We vary the value of $\alpha$ over the range $\left \{0, 0.05, 0.1, 0.15, 0.2, 0.25\right \}$. By observing Table~\ref{tab_alpha}, we can note that a too large $\alpha$ (meaning higher threshold) may keep too much noise and leads to inferior performance. And a too small $\alpha$ will discard many images that can help the training of the segmentation network. According to Table~\ref{tab_alpha}, we empirically set $\alpha = 0.1$.

For the progressive pixel-level noise detection module, we conduct experiments to study the effect of  hyper-parameters of $T_{h}$ and $T_{l}$ that controls the highest and lowest value of the threshold. As shown in Table~\ref{tab_ps}, we first vary the value of $T_{h}$ over the range $\left \{1.0, 1.2, 1.4, 1.6\right \}$ while fixing $T_{l} = 0.8$. After choosing $T_{h} = 1.2$, we vary $T_{l}$ over the range of $\left \{0.6,0.8,1.0\right \}$. By observing Table~\ref{tab_ps}, we can note that a too large $T_{h}$ will keep too much noisy pixels and may not improve the performance very much. In contrast,  a too small $T_{h}$ will also remove many hard pixels that are useful to learn a segmentation network. The network achieves better results when $T_{h}$ is between 1.2 and 1.4. Similarly, a too small or large $T_{l}$ may not facilitate the training process significantly. According to Table~\ref{tab_ps}, we set $T_{h} = 1.2$ and $T_{l}=0.8$ to get the best result of 69.0\%.


For adversarial learning, we conduct experiments to study the effect of hyper-parameter $\lambda_{adv}$ that controls the relative importance of the adversarial loss. By observing Table~\ref{tab_adv}, we can note that a too large $\lambda_{adv}$ may not facilitate the training process. And a too small $\lambda_{adv}$ will dilute the effect of adversarial learning. The network achieves better results when $\lambda_{adv}$ is between 0.0005 and 0.005. According to Table~\ref{tab_adv}, we empirically set $\lambda_{adv} = 0.001$ to get the best result of 69.0\%.

\section{Conclusions}
\label{conclusion}

In this work, we proposed an end-to-end (single-stage) multi-granularity denoising and bidirectional alignment (MDBA) method to address the WSSS task with saliency-generated pseudo labels. Specifically, we proposed an online noise filtering module with class adaptive thresholds and a progressive noise detection module to tackle image-level and pixel-level noise. Moreover, a bidirectional alignment mechanism was proposed to reduce the data distribution gap between simple and complex images at both input and output space with simple-to-complex image synthesis and complex-to-simple adversarial learning. Extensive experiments on PASCAL VOC 2012 and COCO datasets demonstrated the superiority of our approach.

\bibliographystyle{IEEEtran}
\bibliography{egbib}

\end{document}